\pdfoutput=1

\documentclass[11pt]{article}

\usepackage[preprint]{acl}

\usepackage{times}
\usepackage{latexsym}
\usepackage{enumitem}
\usepackage[T1]{fontenc}

\usepackage[utf8]{inputenc}

\usepackage{microtype}
\usepackage{tcolorbox}
\usepackage{inconsolata}
\usepackage{booktabs}
\usepackage{tabularx}
\usepackage{graphicx}
\usepackage{multirow}

\usepackage{makecell}

\usepackage{fontawesome5}
\usepackage{hyperref}

\title{From Babbling to Fluency: Evaluating the Evolution of Language Models\\in Terms of Human Language Acquisition}

\author{
    Qiyuan Yang$^{1}$$^{*}$, Pengda Wang$^{2}$$^{*}$, Luke D. Plonsky$^{3}$, Frederick L. Oswald$^{2}$, and Hanjie Chen$^{1}$ \\
    $^1$Department of Computer Science, Rice University \\
    $^2$Department of Psychological Sciences, Rice University\\
    $^3$Department of English-Applied Linguistics, Northern Arizona University\\ 
    \texttt{\{qy28,pw32,fo3,hc86\}@rice.edu;}
    \texttt{luke.plonsky@nau.edu} \\ 
}

\begin{document}
\maketitle
\begin{abstract}

We examine the language capabilities of language models (LMs) from the critical perspective of human language acquisition. 
Building on classical language development theories, we propose a three-stage framework to assess the abilities of LMs, ranging from preliminary word understanding to complex grammar and complex logical reasoning.\footnote{Code and dataset are available at \url{https://github.com/ericyang1029/Language-Acquisition}}
Using this framework, we evaluate the generative capacities of LMs using methods from linguistic research. 
Results indicate that although recent LMs outperform earlier models in overall performance, their developmental trajectory does not strictly follow the path of human language acquisition. 
Notably, in generation tasks, LMs are more similar to human performance in areas where information is easier to extract from the corpus, such as average word length, clauses, and auxiliary verbs. 
Newer LMs did not exhibit significant progress in terms of specific dimensions, such as clauses and auxiliary verbs, where the variation across corpora is relatively limited. 
Register theory offers a plausible explanation for these observations, suggesting that the linguistic features of the training data have a substantial impact on the models' abilities.
\end{abstract}

\def\thefootnote{*}\makeatletter\def\Hy@Warning#1{}\makeatother\footnotetext{Equal contribution}

\section{Introduction}

Since the advent of early natural language processing (NLP) systems such as ELIZA \citep{weizenbaum1966eliza} and SHRDLU \citep{winograd1971procedures} in the 1950s, researchers have been striving to develop computer programs to understand human language. 
With continuous technological advancements, we have witnessed the rise of language models (LMs), which have achieved unprecedented success in language understanding and language generation (e.g., Gemini, \citealp{team2023gemini}; GPT-4, \citealp{achiam2023gpt}; Llama 3, \citealp{dubey2024llama}). 
These models not only handle complex contexts and generate coherent, human-like text; they also exhibit emergent reasoning abilities and a plausible degree of creativity.

\begin{figure}
    \hspace*{-0.027\linewidth}
    \includegraphics[width=\linewidth]{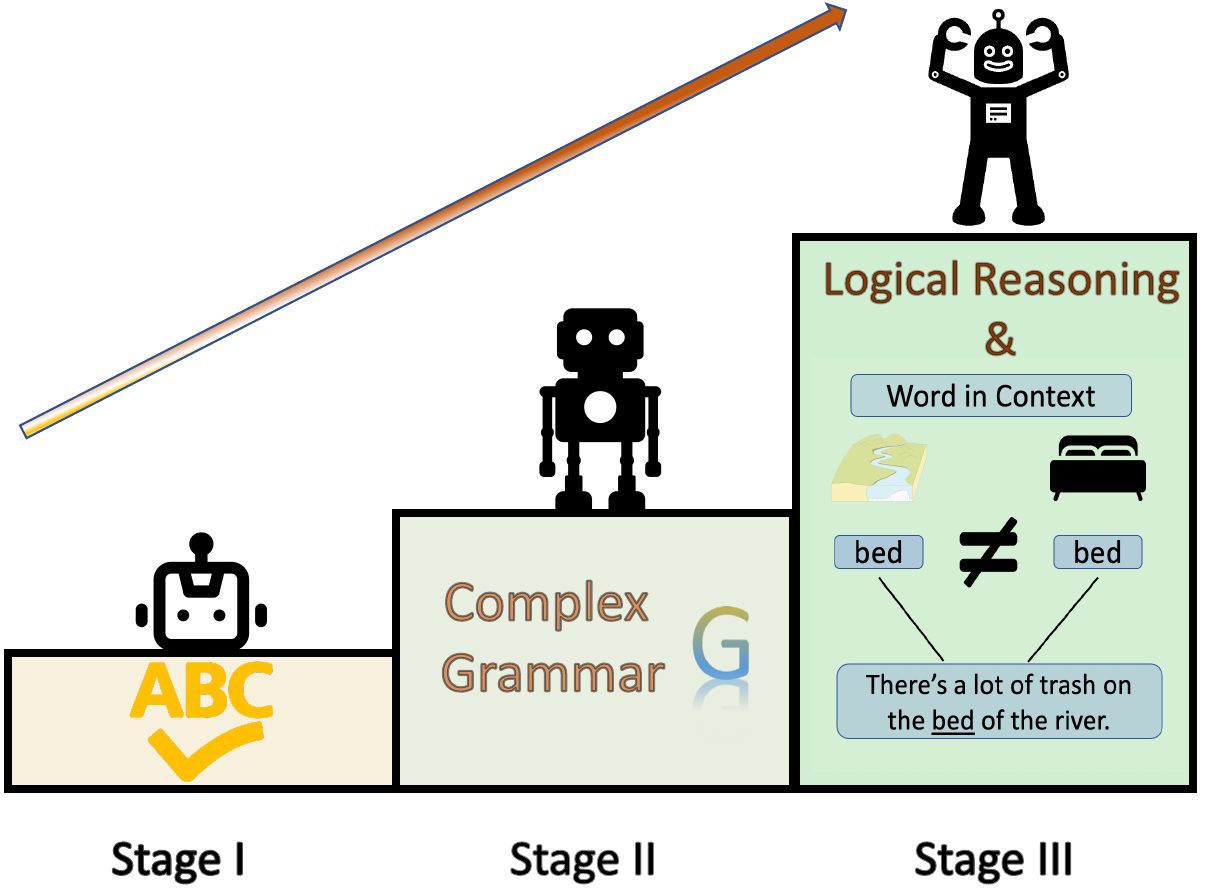}
    \caption{Three-Stage Anatomy of Language Acquisition.}
    \label{fig:teaser}
\end{figure}

As the capabilities of LMs continue to grow, so does the need for comprehensive evaluations of their performance. 
To date, this need has produced a series of benchmark studies that evaluate the capabilities of LMs across various language tasks, such as text classification \citep{sun2023text}, natural language inference (NLI) \citep{ravichander2019equate}, and question answering \citep{kwiatkowski2019natural}, with the goal of comparing different models, identifying their limitations in terms of these tasks, and providing guidance for future model development. 
However, most existing benchmarks, such as GLUE \citep{wang2019gluemultitaskbenchmarkanalysis}, SuperGLUE \citep{wang2020supergluestickierbenchmarkgeneralpurpose} and MMLU \citep{hendrycks2021measuringmassivemultitasklanguage}, while thoroughly evaluating models on specific language tasks, overlook the focus of our current paper: i.e., understanding model capabilities in terms of the developmental stages of human language acquisition \citep{goldberg2009constructions}. 
Similar to how humans acquire language through extensive exposure to spoken or written words as they develop, LMs are similarly trained on large collections of text. 
Both humans and LMs build their language skills by repeatedly encountering language, gradually forming and refining stable patterns and associations. 
Insights from previous studies on the stages of human language development could offer valuable reference points for understanding this process in terms of LMs.

As one of the unique abilities of humans, the acquisition of language has long been a key area of research in psycholinguistics. 
During the process of language acquisition, humans go through multiple stages, from imitation and rule learning to complex contextual understanding \citep{goldberg2009constructions}. 
These stages bear some resemblance to the way current LMs are trained. 
For instance, LMs learn the statistical patterns and grammatical rules of language through training on large-scale data, similar to how infants develop language abilities by receiving a vast amount of input through listening and speaking. 
If we design theory-driven tests based on the human language acquisition process to evaluate the capabilities of LMs, it could help us better understand the nature, potential, and limitations of LMs in their development. 

Our work draws on classical theories of human language development to assess LMs in terms of a three-stage human language development framework \citep{chomsky2014aspects,loban1976language,pinker2003language}, as shown in Figure \ref{fig:teaser}.
The first stage involves developing basic language understanding, similar to early language acquisition in infants. 
At this stage, we evaluate the model's ability to recognize vocabulary, grasp syntax, and perform simple reasoning.
In the second stage, the focus shifts to mastering complex grammar and semantics, where the model demonstrates a deeper understanding of language rules and logical relationships between sentences.
The third stage assesses advanced language abilities, evaluating the model's capacity for complex reasoning and logical analysis.

We further investigate another theory: register theory in linguistics, which posits that different language use scenarios influence the form and structure of language \citep{halliday1977text,matthiessen1993register}. 
This theory offers insights into the extent to which models' abilities depend on the linguistic features encountered in specific situations, referred to as registers. 
In LMs, the training corpus will reflect some registers but not others, which can raise general questions or concerns about the generalizability and biases contained in any given corpus.

We evaluated 15 LMs from 2019 to 2024, excluding instruction fine-tuned or chat versions, with varying parameter sizes (see §\ref{subsec:models}). 
Our findings include: (1) newer LMs generally outperform older ones, though performance varies by task; (2) LMs do not follow human language acquisition patterns but rather reflect changes in architecture and training data; (3) for easily accessible information such as average word length, clauses, and auxiliary verbs, LMs show little improvement over time. 
Recent models have demonstrated minimal progress in these areas due to limited variation across corpora. 
Overall, register theory, which focuses on data, better explains model differences than human developmental processes.

\section{Related Works}
\label{subsec:exResearch}

LMs are computational systems designed to understand and generate text in human language. 
Over time, advancements in LMs, particularly in pre-trained models like GPT \citep{radford2019language} and BERT \citep{devlin2019bertpretrainingdeepbidirectional}, have significantly improved performance across various NLP tasks. 
Large LMs, which leverage vast amounts of data and computational power, can capture more intricate nuances in language \citep{bommasani2022opportunitiesrisksfoundationmodels, wei2022emergentabilitieslargelanguage}, improving its generative capabilities involving masked token or next-token predictions. 

These models are typically fine-tuned for specific tasks after pre-training, further enhancing their adaptability and versatility in practical applications \citep{gururangan-etal-2020-dont}.
As noted previously, systematically evaluating the performance of LMs is critical as they grow in their complexity and diversity \citep{srivastava2023imitationgamequantifyingextrapolating}.
Benchmarking not only provides a standardized way to compare different models; it also highlights areas where improvements are needed, guiding future advancements in the field.

There are many benchmarks that evaluate LMs' abilities. 
Some focus on specific aspects, whereas others cover a broad range of tasks. 
For instance, the SST2 dataset \citep{socher-etal-2013-recursive} measures text classification and the TriviaQA dataset \citep{joshi2017triviaqalargescaledistantly} focuses on question answering. 
Additionally, comprehensive benchmark suites like GLUE \citep{wang2019gluemultitaskbenchmarkanalysis}, SuperGLUE \citep{wang2020supergluestickierbenchmarkgeneralpurpose}, and MMLU \citep{hendrycks2021measuringmassivemultitasklanguage} assess multitask language understanding across a wide range of topics and tasks.
However, these benchmarks do not provide insights about a model's capabilities in terms of human language acquisition, such as in the three-stage framework we provided. 
Insights from previous studies on the stages of human language development may offer valuable reference points for evaluating models' performance.

Previous studies have demonstrated that models can learn hierarchical syntactic structures and exhibit sensitivity to various linguistic phenomena, even when trained with the amount of data that humans typically encounter \citep{milliere2024language, wilcox2024bigger}. 
Assessing these models through the lens of human language development can provide further insights and deepen our understanding of LMs' capabilities.

Human language development is a gradual, stage-based process. 
In the following section (§\ref{subsec:psychlinguisticV}), we will provide a more detailed description of this process, along with a breakdown of language capabilities at each developmental stage.

\section{Psycholinguistics View Framework and Datasets}
\label{subsec:psychlinguisticV}

Psycholinguistics explores the cognitive processes behind language acquisition, focusing on how humans gradually develop language abilities. 
We primarily focus on research related to the various stages of language development.

Previous research has established that language development follows a relatively stable trajectory, with several key stages identifiable along the way. 
For example, \citet{gesell1946child} found that the development of spoken language demonstrates consistent growth, as reflected in metrics such as the average number of words per communication unit, the number of clauses per unit, and the elaboration between subjects and verbs.

Similarly, \citeposs{templin1957certain} analysis of subordinate clause usage also underscores these stages, showing that eight-year-old children use subordinate clauses significantly more often than three-year-olds, marking a pivotal point in language acquisition. 
And \citet{gesell1946child} indicated that the development of spoken language shows a relatively stable growth trend. 
For example, the average number of words per communication unit (C-Unit), the number of clauses in each communication unit, and the amount of elaboration between subjects and verbs all continue to increase. 

\subsection{Framework}
Combining the findings above with those of \citet{watts1944language, o1967syntax, languagedisor} and the summary of \citet{loban1976language}, we can roughly divide the overall process of language development into three stages:

\paragraph{Stage I (Ages 0-6):}
At this stage, children primarily focus on understanding vocabulary, and simple syntactic structures begin to emerge. 
They gradually learn to use pronouns and verbs and become able to distinguish between the present and past tense. 
Although language expression remains relatively simple at this age, the use of compound sentences increases, especially those that express conditionality and causality. 
Using words like ``why,'' ``because,'' and ``if,'' children begin to engage in preliminary causal reasoning, though this ability is not yet fully developed.

\paragraph{Stage II (Ages 6-12):}
During this stage, the development of language gradually moves towards more complex grammatical structures. 
They begin to master finer syntactic elements, such as predicate-argument structures, prepositional phrases, subordinate clauses, and the use of active and passive voice. 
Their semantic understanding also advances, enabling them to grasp the implied meanings of words (e.g., ``run'' implies ``movement'') and handling negation through pre-pending or appending particles to the stem of a word (i.e., morphological negation, refers to the process of creating a negative form of a word by adding a prefix, such as when ``possible'' becomes ``impossible.'' This involves using prefixes like ``un-,'' ``in-,'' or ``im-'' to change the meaning of the original word to its opposite). 

In addition, during this stage, children develop the ability to recognize named entities, quantifiers, and complex concepts such as factuality, symmetry, and redundancy.

\paragraph{Stage III (Above age 12):}
At this stage, children's language abilities are reflected not only in the complexity of their verbal expression but, more significantly, in their use of logical reasoning and abstract thinking.
They begin to engage in spatial reasoning, deductive reasoning, and syllogistic analysis, which allows them to use language with greater precision and rigor. 
Additionally, they become adept at resolving ambiguities in words with multiple meanings and demonstrate a marked improvement in reading comprehension skills.

\subsection{Datasets}
Within each stage we just introduced, we compile several datasets and introduce them in the following section.\footnote{Note that we filter the training dataset and restrict the average C-Unit. In some cases (e.g., bc-if-why), because there is not a sufficient number of filtered examples from its evaluation set, we randomly split off 20\% of the training dataset for validation. For datasets that do not require filtering, the evaluation sets are provided.} 
For an overview of the datasets, please refer to Table \ref{tab:overalldataset} in Appendix \ref{sec:tablegraphs} and see Table \ref{tab:example} in Appendix \ref{sec:tablegraphs} for the example of each dataset.

\subsubsection{Stage I}
\textbf{one-word understanding:} To assess the LM's understanding of individual vocabulary items, we selected examples from publicly accessible vocabulary sample tests \cite{elementary_vocabulary_tests_2024, everyday_vocabulary_collective_nouns_test_2024} and randomly extracted frequently used vocabulary with brief examples from \citet{oxford_learners_dictionaries_2024}.

In this task, LMs will be asked to answer simple multiple-choice questions. 
They will need to choose one of the four choices (a word or phrase) that makes the most sense in the given context.

\textbf{agent-action-object (AAO):} To test whether LMs possess the knowledge to decide whether it is reasonable to take an action on the object, we chose the ``subject-verb-trans'' set from BLiMP \citep{warstadt2023blimpbenchmarklinguisticminimal} as our AAO dataset.

In this task, LMs will be provided two sentences that have minimal differences (one or two words), where one of the two sentences is grammatically correct, and the other is not. LMs will be asked to distinguish between correct and incorrect sentences.

\textbf{bc-if-why:} We select examples containing the words \{because, if, why\} from the Multi-Genre Natural Language Inference (MNLI) dataset \citep{williams2018broadcoveragechallengecorpussentence}, to test the models' preliminary expressiveness in terms of conditionality and causality.

Following the same format in the MNLI dataset, we let the models perform a three-class classification task. Given premise and hypothesis, models will need to classify them into \{entailment, neutral, contradiction\}.

\subsubsection{Stage II}
\textbf{Grammar-comp:} To evaluate complex grammatical structures, we included more comprehensive and diverse grammatical types (e.g. quantifiers, belief verbs) in this task from MNLI \citep{williams2018broadcoveragechallengecorpussentence}. We also exclude instances containing participial words that are not typically mastered at this stage.
We keep the same task setup as in ``bc-if-why'' in Stage I.

\textbf{BLiMP-comp:} To minimize the influence of inference on grammar tasks in addition to MNLI, we extract minimal pair tasks from BLiMP \citep{warstadt2023blimpbenchmarklinguisticminimal}, which includes a wide range of grammatical phenomena, from subject-verb-agreement to syntactic structure. We select those subsets with human average performance of at least 80\% accuracy as tests. The format is the same as the AAO task.

\textbf{CoLA \citep{warstadt2018neural}:} 
Unlike the other two tasks in this stage, models are required to classify a sentence as either grammatically correct or incorrect, assigning it to one of two classes: True or False, respectively.

\subsubsection{Stage III}
\textbf{WiC:} The WiC dataset \citep{pilehvar2019wicwordincontextdatasetevaluating} focuses on words that have multiple meanings. We used it to test the models' ability to probe both the context of the sentences and different definitions of the word when those exist.

In this task, two sentences will be given, where each has one word in common, but they may or may not have the same meanings.
Models will need to judge whether this word has the same meaning or not under these two contexts.

\textbf{ReClor:} This dataset \citep{yu2020reclorreadingcomprehensiondataset} is composed of complex logical reasoning questions. We used it to test whether the models possess complex language abilities, including word understanding, grammatical accuracy, inference, and reasoning.

During this task, models will do multiple-choice questions. Provided with a context and a question, models are expected to choose the most suitable answers to the question from one of four choices.

\section{Experimental Setup}
In this section, we introduce the LMs we tested (§\ref{subsec:models}), the testing methods for different tasks performed by the LMs (§\ref{subsec:tasks}), as well as the evaluation method (§\ref{subsec:evaluation}).

\subsection{Models}
\label{subsec:models}

We investigated 15 LMs in total (excluding instruction fine-tuned or chat versions) over a broad time period (2019 to 2024) and with varying model parameter sizes.

These include GPT-2 (\texttt{gpt-2-large}, \texttt{gpt-2-xl}; \citealp{radford2019language}), RoBERTa (\texttt{RoBERTa-base}, \texttt{RoBERTa-large}; \citealp{liu2019roberta}), ALBERT (\texttt{ALBERT-xlarge}, \texttt{ALBERT-xxlarge}; \citealp{lan2019albert}), Google T5 (\texttt{T5-3b}, \texttt{T5-large}; \citealp{raffel2020t5}), OPT (\texttt{opt-1.3b}, \texttt{opt-2.7b}; \citealp{zhang2022optopenpretrainedtransformer}), Llama2 (\texttt{Llama-2-7b-hf}), Mistral (\texttt{Mistral-7B-v0.3}; \citealp{jiang2023mistral7b}), Llama3 8B (\texttt{Meta-Llama-3-8B}), and Gemma2 (\texttt{gemma-2-2b}, \texttt{gemma-2-9b}).

\subsection{Testing Methods}
\label{subsec:tasks}
We use three different strategies to test the performance of LMs because of the specific task design of certain datasets (e.g., classification), and LMs' architecture differences. 

\textbf{Classification Task}: 
In this type of task, sentences are given as inputs to models. Models will output a class label (e.g., \{0, 1\} for two-class classification, \{0, 1, 2\} for three-class classification).

\textbf{Minimal Pair Task and Vocabulary Task}:
In these two kinds of tasks, we will either calculate the loss for decoder-only models or compare the probability distributions of the masked token through Masked Token Prediction (MLM) (BERT-style) or Span Predictions (T5). Please refer to Appendix \ref{sec:implementation} for details on the format.

\textbf{Reading Comprehension Task}: For this task, we select either the available chat versions or the instruction-fine-tuned versions of our chosen models, as these can be prompted to answer questions in a designated format. 
In addition to the normal prompt, we also apply the zero-shot CoT \citep{wei2022chain} and one-shot ICL \citep{NEURIPS2020_1457c0d6} to determine whether any further improvement in the performance of the LMs can be obtained.

\textbf{Generation Task}: 
The chat and instruction-fine-tuned versions of the models are prompted with instructions for ten different topics, taken from GRE public issue writing prompts \citep{ets_gre}. 
Sample essays with full scores are sourced from \citep{yugt_gre_writing} to compare with the performance of the LMs on this task.

\subsection{Evaluation Method}
\label{subsec:evaluation}

We report accuracy as our main performance metric, as in the original formulation, because most of our testing data is balanced. 
CoLA dataset \citep{warstadt2018neural} also uses the Matthews correlation coefficient (see \ref{formula:mcc}).

\textbf{Normalized Accuracy}: 
Although the NLI task has a baseline accuracy of 0.33 (random guess), tasks with four choices, such as one-word understanding, have a baseline accuracy of 0.25.
Therefore, it is unreasonable to compare them solely on their original accuracy. 
We have therefore normalized each metric by the following formula:\\
\[Normalized\_Accuracy = \frac{A - R}{1 - R}\]
where $A$ is the observed accuracy, $R$ is the accuracy of a random guess. 
This formula is the same as Cohen's kappa coefficient for rating tasks, which takes random rater agreement into account \citep{cohen1960coefficient}.

\section{Experimental Results}

We first analyzed whether the LMs' overall developmental trends between the years 2019 and 2024 were consistent with the developmental trajectory of human language (§\ref{subsec:time}). 
On this basis, we further explored three core questions: 
(1) Did scale matter? 
(2) Did architecture matter? 
(3) Did data matter? 
Finally, we conducted a comprehensive and in-depth evaluation of the models' generative abilities from a linguistic perspective (\S\ref{subsec:generation}).

\begin{figure*}
    \includegraphics[width=\linewidth]{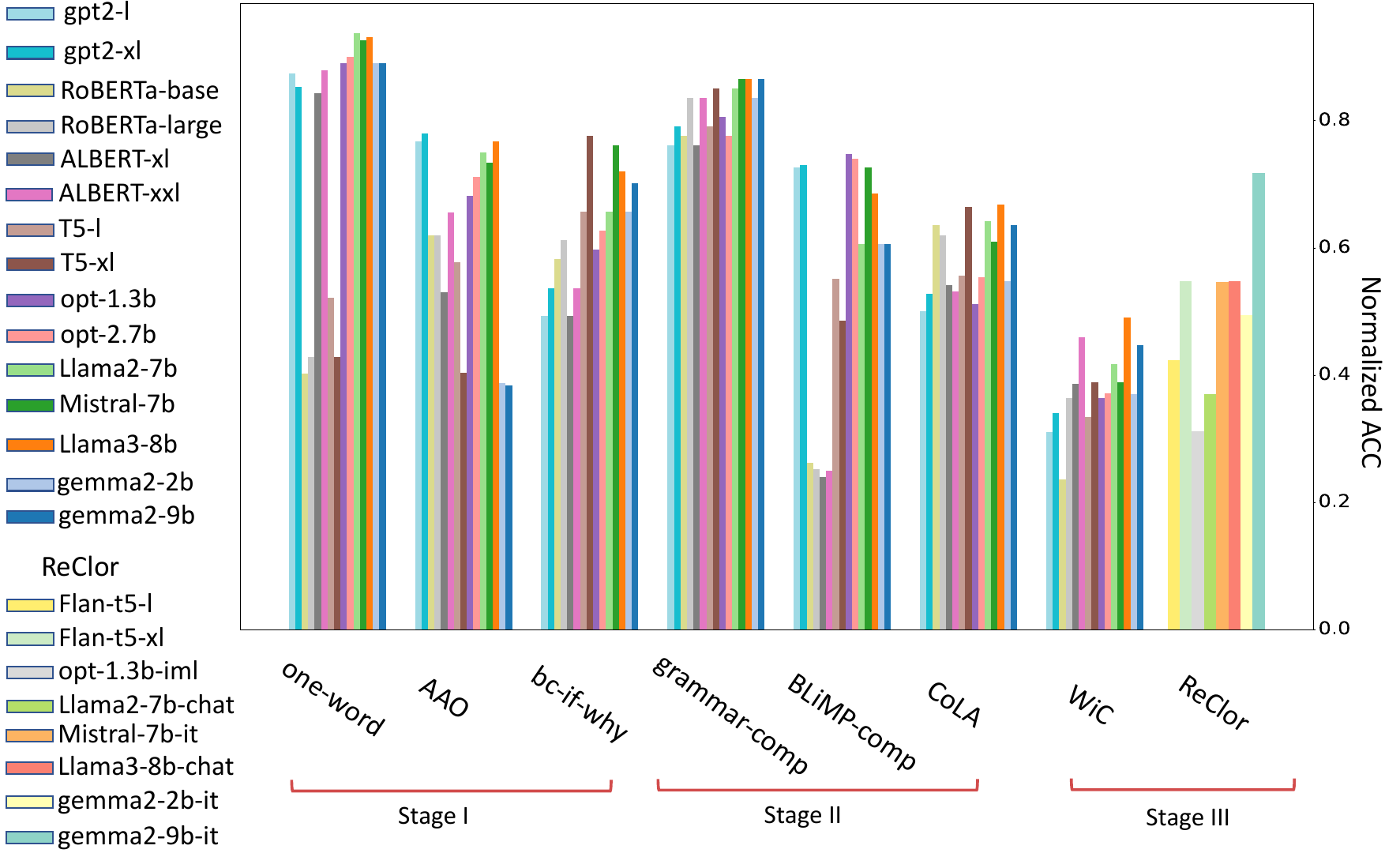}
    \caption{Performance of LMs across three stages. The upper right legend corresponds to models tested in tasks except for ReClor. The lower right legend corresponds to models tested in ReClor. For each task, models are ordered by their time released, and the tie is broken by their parameter sizes. Results from CoLA also use a different metric; please refer to Figure \ref{fig:cola} in Appendix \ref{sec:tablegraphs}.}
    \label{fig:overallgraph}
\end{figure*}

\subsection{Overall Trends in Language Models’ Development}
\label{subsec:time}

Here, we focused on the overall development trends of LMs, and whether these models mimic the process of human language acquisition.
As noted previously, just as humans learn language from an early age by being exposed to a large amount of spoken or written language, LMs are trained on vast text corpora. 
Both humans and LMs develop language abilities through repeated exposure to language, forming patterns and associations over time. 
Previous research on the stages of human language development may serve as a reference.

As mentioned earlier, these datasets have been divided into tasks based on theories of human language development. 
We anticipated that certain LMs would exhibit stronger performance in the early stages of language acquisition but show more modest results in the later stages.
Further, if these stages of human language development hold for the development of LMs, then if an LM achieves relatively good results in the third stage, then it should also demonstrate corresponding success in the first and second stages on which the third stage depends. 
Despite this theoretical motivation, the experimental results did not support this hypothesis.

Figure \ref{fig:overallgraph} displays our overall results.
In Stage I, we first tackled fundamental tasks of human language acquisition, such as understanding individual words. 
Most models performed well at this stage, but a few lagged behind. 
For example, the accuracy of T5 and RoBERTa was only half that of other models in one-word understanding. 
We found that Gemma2 performed well in many tasks; however, it fell short compared to other models on the AAO task. 
After conducting some experiments (see Appendix \ref{sec:casestudy}) on these models, we discovered that T5 and RoBERTa did not perform well on questions requiring contextual information. 
However, the fine-tuned versions of T5 and Gemma2 excelled in one-word understanding and the AAO task, respectively.

Stage II involved more complex grammatical knowledge, yet most LMs did not share this difficulty, performing as well as, or even better than, they did in stage I.
Notably, despite similar overall performance, there were significant differences in the models' scores across different grammatical phenomena from BLiMP-comp. Please refer to Table \ref{tab:blimp-comp} in Appendix \ref{sec:tablegraphs} for detailed examples.

In Stage III, performance differences among the LMs became more pronounced across various tasks.
For the WiC task, the LMs failed to demonstrate comparative performance relative to other tasks in Stage I and Stage II.
In the ReClor task, the fine-tuned opt-1.3b model and Llama2-chat version performed poorly, while Gemma2-9b-instruct achieved higher accuracy. 
Moreover, one-shot ICL and CoT learning did not significantly improve model performance in this task (see Table \ref{tab:reclorfull} in Appendix \ref{sec:tablegraphs} for more details).

\paragraph{Does Scale Matter?}

Although previous research has shown that the performance of LMs often improves with the expansion of model parameters \citep{kaplan2020scalinglawsneurallanguage}, in most of the ability tests we conducted across different stages of language development, there was no significant difference in performance between small models and their larger counterparts. 
The only exception was the complex task ReClor (in Stage III), where larger models significantly outperformed smaller ones.

Just like previous research (e.g., \citealp{milliere2024language, wilcox2024bigger}), our results also support the idea that small models can effectively encode sufficient information for certain tasks, meaning that increasing model parameters is not the only path to improving performance.
Therefore, instead of solely pursuing larger models, drawing insights from linguistic research might be a more effective way to enhance overall model performance \citep{milliere2024language, wilcox2024bigger}.

\paragraph{Does Architecture Matter?}
\label{ref:archi}
We noticed that, in classification tasks, encoder models (including T5, which only uses its encoder part for classification), even with smaller numbers of parameters, almost equalize or exceed the performances of decoder models with larger numbers of parameters. 
The bidirectional property of encoder models could contribute to this. 

To master NLI and WiC tasks, it is pivotal to possess the inter-relationship between tokens in two sentences. 
Consequently, models with encoders could cross-attend to previous and later contextual information in the question and thus manage such tasks well.

For tasks that compare loss between sentence pairs (AAO and one-word), most decoder-only models, such as GPT-2, outperform encoder-only or encoder-decoder models (e.g., T5 and RoBERTa).
The differences in architecture determine how they tackle such problems, particularly with prediction loss (e.g., MLM vs. next-token prediction). 

We suspect that the randomness introduced by masking tokens (or corruption rates for T5) could contribute to this difference. 
Additionally, Next Sentence Prediction (NSP) might play an important role in one-word understanding tasks. 
Even with larger batch sizes, models such as RoBERTa and T5, which are not trained on NSP, may lack the ability to model sentence-to-sentence transitions, which is essential for that task.

\paragraph{Do Data Matter?}
As the representations in AI models are converging \cite{huh2024platonicrepresentationhypothesis}, the scale and the quality of data that they learn from are the key to their performance. 
We found that as models' pretraining data scale up, regardless that bigger is not always better, there was a trend to perform better in each stage (see Figure  \ref{fig:stageIdata}, \ref{fig:stageIIdata}, \ref{fig:stageIIIdata} in Appendix \ref{sec:tablegraphs}). 

Noticeably, Mistral keeps an impressive performance--to-data volume ratio, but it does not bear this advantage in stage III. 
Although there might be disparities among model sizes, we could anticipate that with a larger amount of training data, LMs could learn richer knowledge and generalize it better.

\subsection{Language Models' Generation Ability}
\label{subsec:generation}

We also evaluated the generation abilities of some LMs through the generation task. 
Here, we regard generation ability as a reflection of LMs' overall capability, as generation requires word-level understanding, flexible use of grammatical knowledge, and strong logical reasoning skills to ensure sentence completeness and fluency.

In the field of linguistics, extensive research has explored co-occurrence patterns of language features. 
Drawing on the study of the Multi Dimensional Analysis Tagger (MAT) by \citet{nini2019}, which replicates the procedure by \citet{biber1988}, we compared five representative dimensions.

\textbf{NN} (nouns that are not identified as nominalizations or gerunds): The use of nouns is an important component of syntactic structure, helping to assess whether the model handles nouns accurately and flexibly.

\textbf{AWL} (average word length): Word length reflects the complexity of the generated text and the diversity of language style, measuring the model's lexical richness.

\textbf{Clause} (a collection of adjectival and adverbial clauses): The frequency and diversity of clause use reflect the model's ability to generate complex sentences, showcasing its mastery of advanced grammar.

\textbf{TTR} (type-token ratio): This dimension evaluates the richness of the generated text in terms of lexical diversity, indicating the model’s flexibility in word choice.

\textbf{Auxiliary verbs} (e.g., modal verbs expressing possibility, prediction, and necessity): The use of auxiliary verbs reveals whether the model can express complex reasoning and logical relationships, serving as an important indicator of reasoning ability in generation tasks.

\begin{figure}[h]
    \centering
    \includegraphics[width=\linewidth]{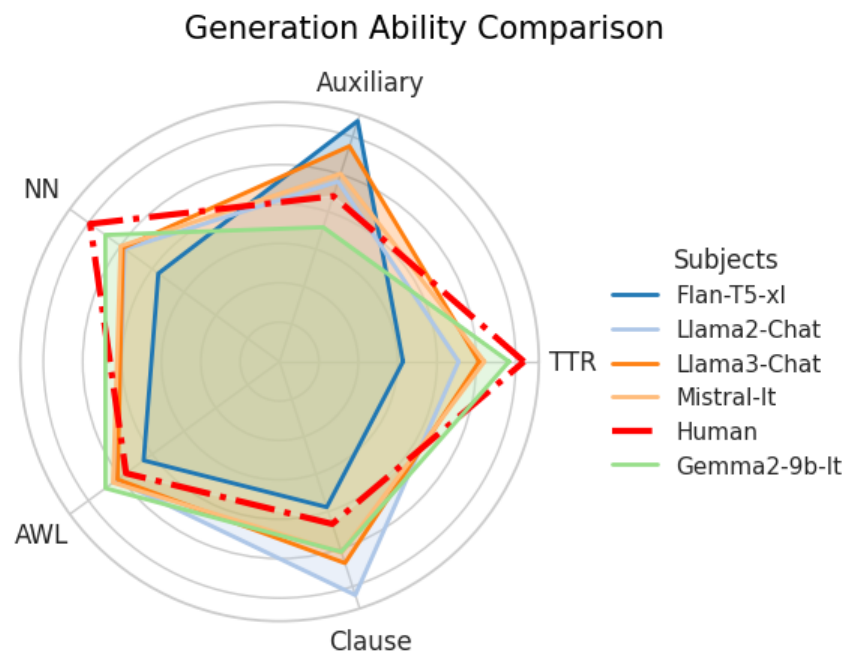}
    \caption{Generation Abilities of six models along five selected dimensions. }
    \label{fig:generationradar}
\end{figure}

In all five dimensions, we found that humans tend to exhibit more variation than LMs in the NN (noun usage) and TTR (type-token ratio) dimensions, whereas no significant differences were observed on the other three dimensions (see Figure \ref{fig:generationradar}). We believe this variation is due to the following reasons:

(1) NN: Humans, when using language, tend to flexibly choose vocabulary and expressions depending on the context, topic, and purpose of communication. 
The use of nouns may reflect humans’ ability to name objects, concepts, or abstract ideas within a specific context, and this ability becomes more diverse as topics change. 
On the other hand, LMs, though trained on large corpora, may rely on more frequent patterns or words during generation rather than adjusting as flexibly as humans do based on context.

(2) TTR: Human language ability is often characterized by broad vocabulary use, especially when dealing with complex or rich topics, where such lexical diversity becomes more apparent. 
In contrast, LMs might tend to use more common words when generating text, particularly if certain words appear more frequently in the training data, leading to less flexibility in lexical diversity, compared to humans.

Overall, although LMs can simulate a certain level of human linguistic diversity, we believe that due to their reliance on training data, they may not exhibit the same level of variation and flexibility as humans when producing new linguistic expressions. 
In the other three dimensions (such as word length, use of subordinate clauses, and auxiliary verbs), the differences between LMs and humans were smaller, likely because these dimensions depend more on grammatical structure and syntactic rules, which are more clearly defined in the training of LMs, allowing them to match human performance in these areas.

\paragraph{Language Models’ Development in Generation}
We also explored the relationship between these five dimensions and the development trends of LMs. 
We found that except for Clause and Auxiliary verbs, NN, AWL, and TTR showed significant progress (see Figure \ref{fig:generation}).
This phenomenon may be due to improvements in the training corpora for models. 
The progress in NN and AWL may reflect an enhancement in the models' ability to generate complex and precise vocabulary. 
As LMs developed, their vocabulary size, semantic understanding, and contextual processing capabilities improved through learning from training data, enabling them to generate richer vocabulary and longer, more complex structures.
The increase in TTR indicates that the model can use a wider range of vocabulary when generating text, rather than repeatedly using the same words. 
This could be attributed to the model’s ability to better capture lexical diversity when processing large-scale training corpora and reflect this diversity in its generation tasks.

In contrast, the trends for Clause and Auxiliary verbs showed less noticeable changes, possibly because these features involve more complex grammatical structures and logical reasoning. 
Models have made progress in vocabulary generation, yet they still face significant challenges in accurately generating more complex clauses and auxiliary verbs. 
This may require deeper syntactic understanding and stronger logical reasoning abilities, which are improving at a slower pace.

\section{Conclusion}
We evaluated LMs by incorporating theories from human language acquisition. 
Building on classical language development theories, we proposed a three-stage framework to assess the abilities of LMs. 
By and large, we observed that LMs do not conform to human language acquisition patterns. 
Although some LMs performed competitively in the later stages, they struggled with tasks in the earlier stages. 
This may be due to their specific architectures, parameter sizes, and the language corpora they were trained on.

Models show smaller differences from human performance in areas where information is easier to extract from the corpus, such as average word length, clause structure, and auxiliary verb usage in generation tasks. 
For dimensions that do not vary significantly across corpora, the models' performance similarly does not show a significant improvement. 

Register theory offers a plausible explanation for these observations, suggesting that the linguistic features of the training data substantially influence the models' abilities.

\section*{Limitations}
This evaluation was necessarily limited by the genres of our collected dataset, which consisted entirely of text.
Texts represent only part of the information acquired during human language acquisition. 
For example, \citet{Barreto2019} introduced visual questions in the CELF-5 that assessed children's understanding of spatial terms, requiring the examinee to identify the position of an object in a picture. 
Similarly, the TOLD-P:5 \citep{newcomer2018test} assessed children's spoken language skills through tasks such as defining spoken words and demonstrating an understanding of their meanings.
To explore this topic further, a multimodal dataset incorporating images, videos, and speech would have been necessary.

Moreover, since the aforementioned assessments were commercially available, accessibility issues arose concerning such datasets. 
In the spirit of open science, future work should focus on creating similar datasets that are open to a wide range of research communities.

Additionally, research by \citet{McMurray2014} showed individual differences in human language abilities. 
Similarly, LMs could have been developed to model such variations more closely.

Finally, due to the rapid advancements in LMs and their increasing parameter sizes, a continuous and sustainable evaluation of these models might have been required.

\section*{Ethics Statement}
The datasets we compiled are all publicly available for research purposes (under CC-BY 4.0 license or unspecified). 
We have manually checked each example from the one-word understanding we collected and modified to ensure it does not contain any harmful information or bias.


\bibliography{acl_latex}

\clearpage
\newpage
\appendix
\section{Appendix A: Case Study}
\label{sec:casestudy}
\begin{tcolorbox}[colback=blue!5!white, colframe=gray!75!black, 
                  title=An example question in one-word-understanding that T5 made a mistake, width=\linewidth, sharp corners]
\textbf{Model choice:} wait \\
\textbf{Correct choice:} rush \\

\textit{You don’t have to \_\_\_\_\_\_! We’re not late!}
\begin{itemize}[itemsep=-3pt]
\item[A)] dream
\item[B)] laugh
\item[C)] rush
\item[D)] wait
\end{itemize}
\label{example:oneowrdmistake}
\end{tcolorbox}

We also investigate questions that RoBERTa and T5 answered incorrectly in the one-word understanding task, which all other models, including decoder-only and encoder-only models, answered correctly. 
After a thorough inspection of the testing examples that RoBERTa and T5 did not answer correctly, we identified two common points:
(1) The models tend to choose answers that form more frequent collocations. 
For example, the models prefer ``think about'' over ``complain about.'' 
``Think about'' can be used in a wider variety of contexts, including contemplation, consideration, and planning, whereas ``complain about'' has a negative connotation and is more context-specific.
(2) Most of these questions require information from the surrounding context, either before or after the blank that needs to be filled in, which is similar to the findings of the case study in \citet{wang2024will}.

We carefully selected 50 examples from our training dataset on one-word understanding and tested RoBERTa-base and T5-large on these examples. 
All of the selected questions are composed of either those requiring context knowledge or those relying solely on collocation knowledge. 
To solve example \ref{example:oneowrdmistake}, the models must attend to the second sentence to understand that ``not late'' is related to ``don't have to rush,'' rather than focusing solely on the first sentence.

\paragraph{RoBERTa}
RoBERTa-base answered 23 out of 50 examples correctly with an accuracy of 46\%.
Upon closer investigation, we found that, out of the 27 questions RoBERTa made mistakes on, 60\% (16 questions) required context, while 40\% (11 questions) were related to collocation.

\paragraph{T5}
For the same set of examples, T5-large correctly answered 28 out of 50 examples, achieving an accuracy of 56\%. 
Of the 22 questions that T5 answered incorrectly, 16 (73\%) required some contextual knowledge, while 6 (27\%) involved collocations.

Since T5 performed relatively well compared to other models, we speculate that the way it handles multiple-choice questions contributes to its lower performance (see §\ref{ref:archi}).  
As a result, we tested Flan-T5 (both \texttt{large} and \texttt{3b}) on this task. We found that their performance, measured by normalized accuracy, increased to 0.807 (Flan-T5-l) and 0.898 (Flan-T5-xl).

\paragraph{Gemma2}
Similarly, we tested instruction-fine-tuned versions of Gemma2 on the AAO task, where it underperforms. Their normalized accuracy rises to 0.87 and 0.85 for the 2b and 9b models, respectively, approaching the performance of other models.
By fine-tuning on a wider variety of datasets, it enables generalization across a range of tasks.

\section{Appendix B: Data Contamination}
There has been an increasing concern in data contamination nowadays \citep{deng2024investigatingdatacontaminationmodern}.
In this section, we investigate whether the pretraining data contain any datasets used in our evaluation.
We apply the MIN-K\% Prob method \cite{shi2024detectingpretrainingdatalarge}. 
This method selects the top k\% of tokens with the highest negative log-likelihood and then computes the average log-likelihood.
It is based on the hypothesis that an unseen example is likely to contain a few outlier words with low probabilities under the LMs, whereas a seen example is less likely to have words with such low probabilities. 
We follow the same settings as in that research and choose $k = 20$. 
If the number of tokens is between zero and one after multiplying the token length by 20\%, we round it up to one.

In the following paragraph, we list the selection methodology:

\textbf{one-word-understanding:} We selected all instances of our test datasets and included sentences containing the correct answers.

\textbf{AAO:} We selected all examples from the test set, including both sentence\_good and sentence\_bad.

\textbf{bc-if-why:} We included all instances in the test datasets, incorporating both the premise and the hypothesis.
   
\textbf{grammar-comp:} In the test data, we randomly selected 1,000 examples and kept all other settings the same as in bc-if-why.
   
\textbf{BLiMP-comp:} For each grammatical phenomenon, we selected 50 examples, resulting in 2,800 instances. All other settings were the same as in AAO.
    
\textbf{CoLA:} All of the test examples were selected.
    
\textbf{grammar-diag:} We included all of the examples in the test datasets. The settings were the same as in bc-if-why.

\textbf{WiC:} Both sentences, one and two, were included.
    
\textbf{ReClor:} We tested the ``context'' part in each question. 
For this question, we tested the instructional fine-tuned and the chat version of the models.

Across each task, we presented the average MIN-K\% probability for all individual sentences. 
For encoder-only models, we adapted this method by calculating the logits after masking each token in every sentence.
To measure the relative MIN-K\% probability, we randomly generated a sequence of all alphabets with a length of 10.

Overall, all models demonstrated comparatively low probabilities. We found that, in most datasets, the models are within 5\% of the probabilities from random letters. 
However, gemma2-2b slightly exceeds 5\% in the AAO dataset, which we consider acceptable (see Table \ref{tab:mink}).

\newcolumntype{C}{>{\centering\arraybackslash}X}
\newcommand{\result}[1]{#1}
\begin{table*}[!t]
\scriptsize  
\renewcommand{\arraystretch}{1.2}  
\begin{tabularx}{\textwidth}{@{} l *{10}{C} @{}}
\toprule
\textbf{Models} & 
\textbf{AAO} & 
\textbf{one-word} & 
\textbf{bc-if-why} & 
\textbf{grammar-comp} & 
\textbf{BLiMP-comp} & 
\textbf{CoLA} & 
\textbf{grammar-diag} & 
\textbf{WiC} & 
\textbf{ReClor} &
\textbf{Random letters}
\\
\midrule
opt-1.3b & 
\result{12.75} & 
\result{10.18} & 
\result{9.13} & 
\result{9.42} & 
\result{12.51} & 
\result{10.37} & 
\result{9.11} & 
\result{10.41} & 
\result{10.30} &
\result{10.29}\\
opt-2.7b & 
\result{12.8} & 
\result{10.17} & 
\result{9.16} & 
\result{9.43} & 
\result{12.54} & 
\result{10.38} & 
\result{9.05} & 
\result{10.39} & 
\result{/} &
\result{10.22}\\
T5-large & 
\result{12.75} & 
\result{10.18} & 
\result{9.13} & 
\result{9.42} & 
\result{12.78} & 
\result{10.37} & 
\result{9.11} & 
\result{10.41} & 
\result{0.73} &
\result{4.88}\\
T5-3b & 
\result{12.75} & 
\result{10.18} & 
\result{9.13} & 
\result{9.42} & 
\result{13.27} & 
\result{10.37} & 
\result{9.11} & 
\result{10.41} & 
\result{0.62} &
\result{5.00}\\
gpt2-large & 
\result{12.66} & 
\result{10.55} & 
\result{8.91} & 
\result{9.15} & 
\result{12.54} & 
\result{10.04} & 
\result{9.02} & 
\result{9.73} & 
\result{/} &
\result{9.87}\\
gpt2-xl & 
\result{12.67} & 
\result{10.47} & 
\result{8.86} & 
\result{9.13} & 
\result{12.38} & 
\result{10.04} & 
\result{8.99} & 
\result{9.70} & 
\result{/} &
\result{9.84}\\
Llama2-7b & 
\result{11.58} & 
\result{9.24} & 
\result{8.96} & 
\result{8.87} & 
\result{11.29} & 
\result{9.63} & 
\result{8.36} & 
\result{9.89} & 
\result{8.31} &
\result{9.87}\\
Llama3-8b & 
\result{13.13} & 
\result{10.35} & 
\result{9.80} & 
\result{9.70} & 
\result{12.69} & 
\result{10.58} & 
\result{9.03} & 
\result{10.85} & 
\result{11.00} &
\result{11.00}\\
Mistral-7b & 
\result{12.16} & 
\result{9.80} & 
\result{9.64} & 
\result{9.42} & 
\result{12.14} & 
\result{10.18} & 
\result{8.72} & 
\result{11.27} & 
\result{7.08} &
\result{10.12}\\
gemma-2-2b & 
\result{20.22} & 
\result{14.06} & 
\result{13.25} & 
\result{13.52} & 
\result{19.60} & 
\result{15.22} & 
\result{12.62} & 
\result{16.26} & 
\result{8.62} &
\result{15.54}\\
gemma-2-9b & 
\result{22.14} & 
\result{14.82} & 
\result{13.85} & 
\result{14.03} & 
\result{21.50} & 
\result{16.12} & 
\result{12.94} & 
\result{16.63} & 
\result{9.11} &
\result{17.11}\\
ALBERT-xlarge & 
\result{11.62} & 
\result{8.72} & 
\result{7.46} & 
\result{7.65} & 
\result{11.03} & 
\result{8.13} & 
\result{7.08} & 
\result{8.27} & 
\result{/} &
\result{11.19}\\
ALBERT-xxlarge & 
\result{12.65} & 
\result{8.72} & 
\result{7.46} & 
\result{7.65} & 
\result{12.07} & 
\result{8.13} & 
\result{7.08} & 
\result{8.27} & 
\result{/} &
\result{11.17}\\
RoBERTa-base & 
\result{12.87} & 
\result{9.29} & 
\result{7.27} & 
\result{7.10} & 
\result{11.92} & 
\result{7.91} & 
\result{5.82} & 
\result{7.99} & 
\result{/} &
\result{9.89}\\
RoBERTa-large & 
\result{12.50} & 
\result{8.81} & 
\result{6.83} & 
\result{6.62} & 
\result{11.50} & 
\result{7.61} & 
\result{5.36} & 
\result{7.45} & 
\result{/} &
\result{9.29}\\
\bottomrule
\end{tabularx}
\caption{MIN-K\% Prob measured in \%. Models measured in the ReClor task are the fine-tuned or chat version of that model.}
\label{tab:mink}
\end{table*}

\section{Appendix C: Implementation Details and Metrics}
\subsection{Implementation Details}
\label{sec:implementation}

\paragraph{Classification}
For BERT-style encoder models \citep{devlin2019bertpretrainingdeepbidirectional}, a special token, \texttt{[CLS]}, is used as input to an MLP for prediction. 
In decoder models such as GPT-2 \citep{radford2019language}, the hidden state of the last token is connected to a classification head. 
For T5 \citep{raffel2020t5}, with an encoder-decoder architecture, we use only the encoder to make predictions. 
Since an MLP is concatenated to each model, fine-tuning is necessary for the models to perform classification. Otherwise, the results will be random guesses. 
We fine-tune the models on grammar-comp for 1 epoch due to the large amount of data, and other classification tasks for 20 epochs maximum using four NVIDIA A-6000 GPUs. 
The learning rates we used range from 1e-6 to 1e-4, depending on model sizes and data sizes. Training batch sizes range from 1 to 16, given different parameter sizes. We also use LoRA \citep{hu2021lora} for models with large parameter sizes (Llama2-7b, Llama3-8b, Mistral-7b, Gemma2-9b) due to the limitations of computational resources.

\paragraph{Minimal Pair and Vocabulary}
For decoder models, the average loss of the sequence is computed to determine which sentence is better. 
For BERT-style models, Masked Language Modeling is used to make predictions. For minimal pair questions (AAO and BLiMP-comp), special masks (e.g., \texttt{<MASK>}) are placed at the positions where the two sentences differ.
Of the masked words, we select the one with a larger probability among the prediction of the masked positions. Similarly, for one-word understanding, we masked the blanks in the sentence. Then we choose one of the four words/phrases with the largest probability.
T5, which is very similar to BERT-style models, uses Span Predictions. We compare the probability of the words it predicts between the span: \texttt{<extra\_id\_0>} word(s) predicted \texttt{<extra\_id\_1>}.

\paragraph{Generation Configuration}
The number of tokens generated by the LMs is set between a minimum of 500 and a maximum of 600 to ensure meaningful and comparable results across all chosen models. 
We keep the default generation parameters for all models, with two exceptions: Flan-T5 \citep{chung2022scalinginstructionfinetunedlanguagemodels} and OPT-IML \citep{iyer2023optimlscalinglanguagemodel} tend to generate repetitive sentences, so we relax their sampling criteria and apply top-k sampling with a probability of 0.9.

\paragraph{Other}
For filtering examples from datasets, we use the nltk \citep{nltk} and spaCy\citep{spacy} packages in Python.

\subsection{Matthews Correlation Coefficient Formulation:}
\label{formula:mcc}
MCC = \\
\begin{equation}
    \small \frac{TP \cdot TN - FP \cdot FN}{\sqrt{(TP + FP)(TP + FN)(TN + FP)(TN + FN)}}
\end{equation}
where:
\begin{itemize}
    \item FP: False Positive
    \item FN: False Negative
    \item TP: True Positive
    \item TN: True Negative
\end{itemize}

\section{Appendix D: Interdisciplinary Collaboration}
We would like to emphasize the importance of interdisciplinary collaboration. 
As LMs continue to evolve and mature, their potential applications across various fields are becoming increasingly evident. 
For example, they can be used in sports assessment \citep{xia2024sportu}, assist in questionnaire design in the social sciences \citep{wangnot2024}, answer clinical case questions \citep{chen2024benchmarking}, and even help with candidate screening \citep{wang2024putting}.

Interdisciplinary collaboration not only provides innovative technological solutions for various fields, but also brings unique insights from different disciplines into computer science, facilitating a better understanding of the underlying problems. 
For instance, collaboration between computer science, linguistics, and psycholinguistics offers new perspectives and methods, aiding in understanding the natural language processing capabilities of models from the viewpoint of language formation and development.

Such interdisciplinary collaboration transcends the limitations of individual disciplines, fostering the integration and innovation of knowledge, and enabling more complex and intelligent technological solutions. 
This trend presents new opportunities for future research and practice, driving societal progress.

\newpage

\section{Appendix E: Tables and Graphs}
\label{sec:tablegraphs}

\begin{table}[h]
\centering
\small
\begin{tabular}{ll|>{\centering\arraybackslash}p{0.5cm}>{\centering\arraybackslash}p{0.5cm}|>{\centering\arraybackslash}p{2cm}>{\centering\arraybackslash}p{1cm}}
\toprule
\multirow{2}{*}{Stage} & \multirow{2}{*}{Type} & \multicolumn{2}{c|}{Data Split} & \multirow{2}{*}{Aspect} \\
& & Train & Test & \\
\midrule
& one-word & 598 & 255 & word-level\\
\multirow{6}{*}{I} 
& & &\\
& \multirow{1}{*}{AAO} & \multirow{1}{*}{-} & \multirow{1}{*}{1k} & preliminary common sense\\
& & &\\
& \multirow{1}{*}{bc-if-why} & \multirow{1}{*}{1.4k} & \multirow{1}{*}{348} & causality conditionality\\
\midrule
\multirow{8}{*}{II} 
& grammar-comp & 170k & 19k & \multirow{8}{*}{grammar}\\
& & &\\
& CoLA & 6.8k & 1.7k\\
& & &\\
& grammar-diag & - & 645 \\
& & &\\
& BLiMP-comp & - & 56k\\
\midrule
& WiC & 5.4k & 1.4k & word meaning under context\\
\multirow{6}{*}{III}
& & &\\
& ReClor & 4.6k & 1k & logical reasoning\\
& & &\\
& generation & - & 10 & logical composition\\
\bottomrule
\end{tabular}
\caption{Tasks from different stages. The Aspect column lists different language aspects tested. AAO = agent-action-object; one-word = one-word understanding dataset.}
\label{tab:overalldataset}
\end{table}

\begin{figure*}[h]
    \centering
    \includegraphics[width=\linewidth]{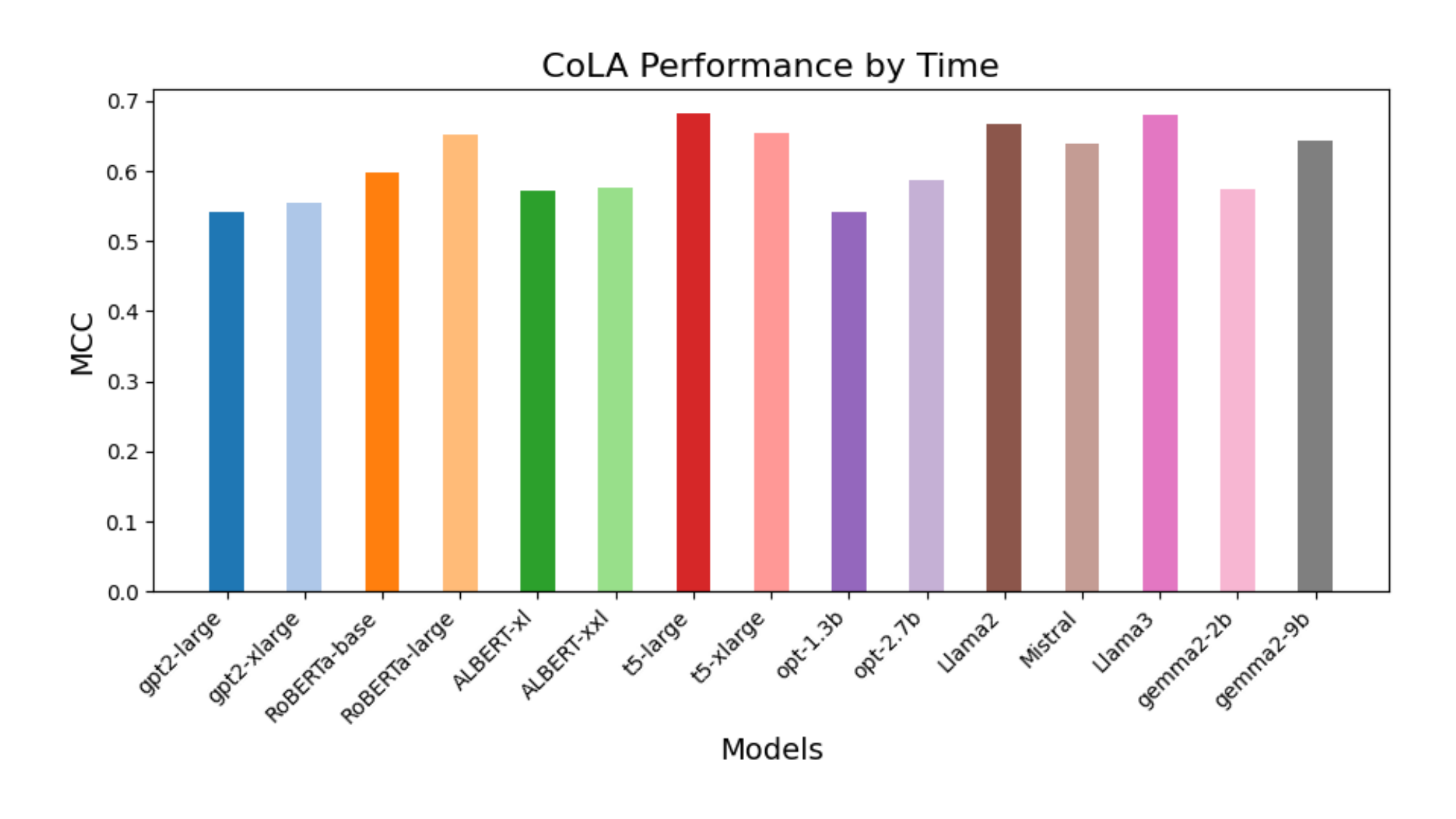}
    \caption{CoLA performance in Stage II measured in Matthews Correlation Coefficient (\ref{formula:mcc}). The result is obtained by training models at most 20 epochs}
    \label{fig:cola}
\end{figure*}

\begin{figure*}[h]
    \centering
    \includegraphics[width=\linewidth]{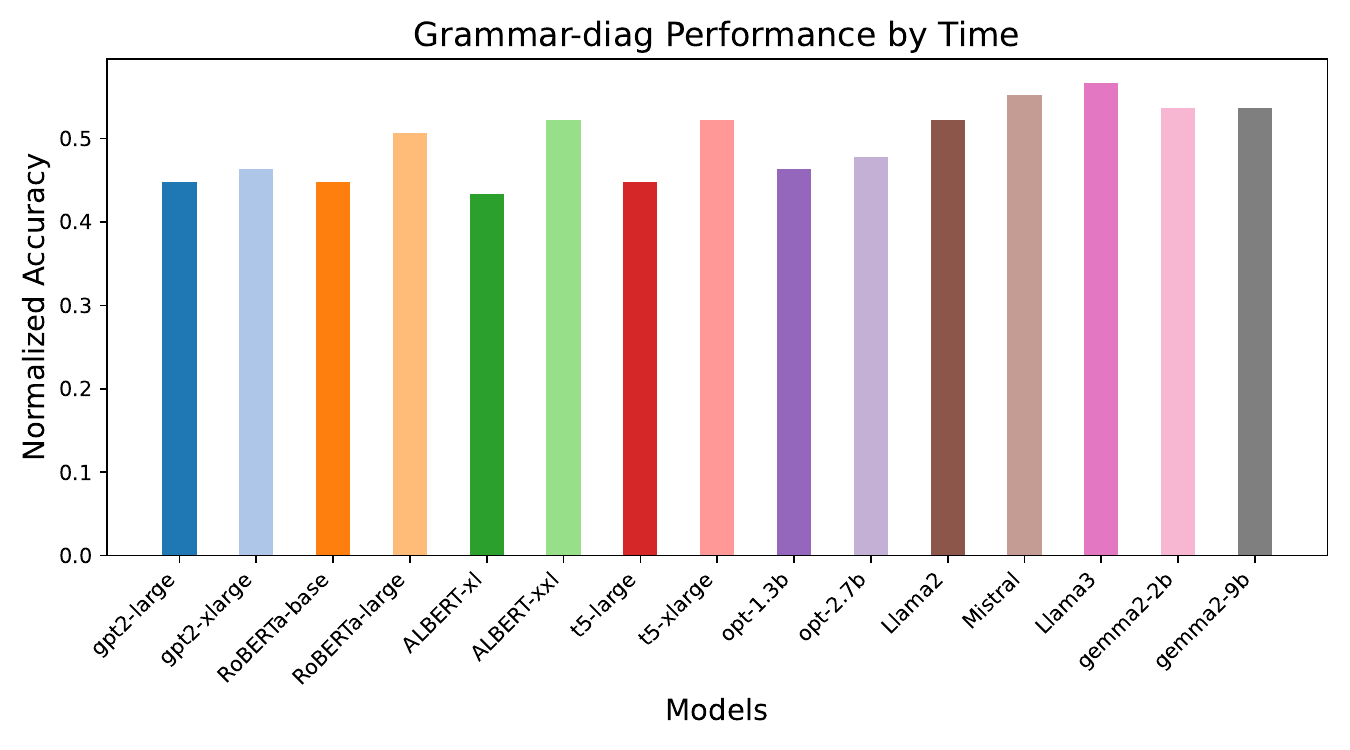}
    \caption{Grammar-diag performance in Stage II. Models are ordered by time. We test on models after fine-tuning on bc-if-why and grammar-comp's training set.}
    \label{fig:grammar_diag}
\end{figure*}

\begin{figure*}[h]
    \centering
    \includegraphics[width=\linewidth]{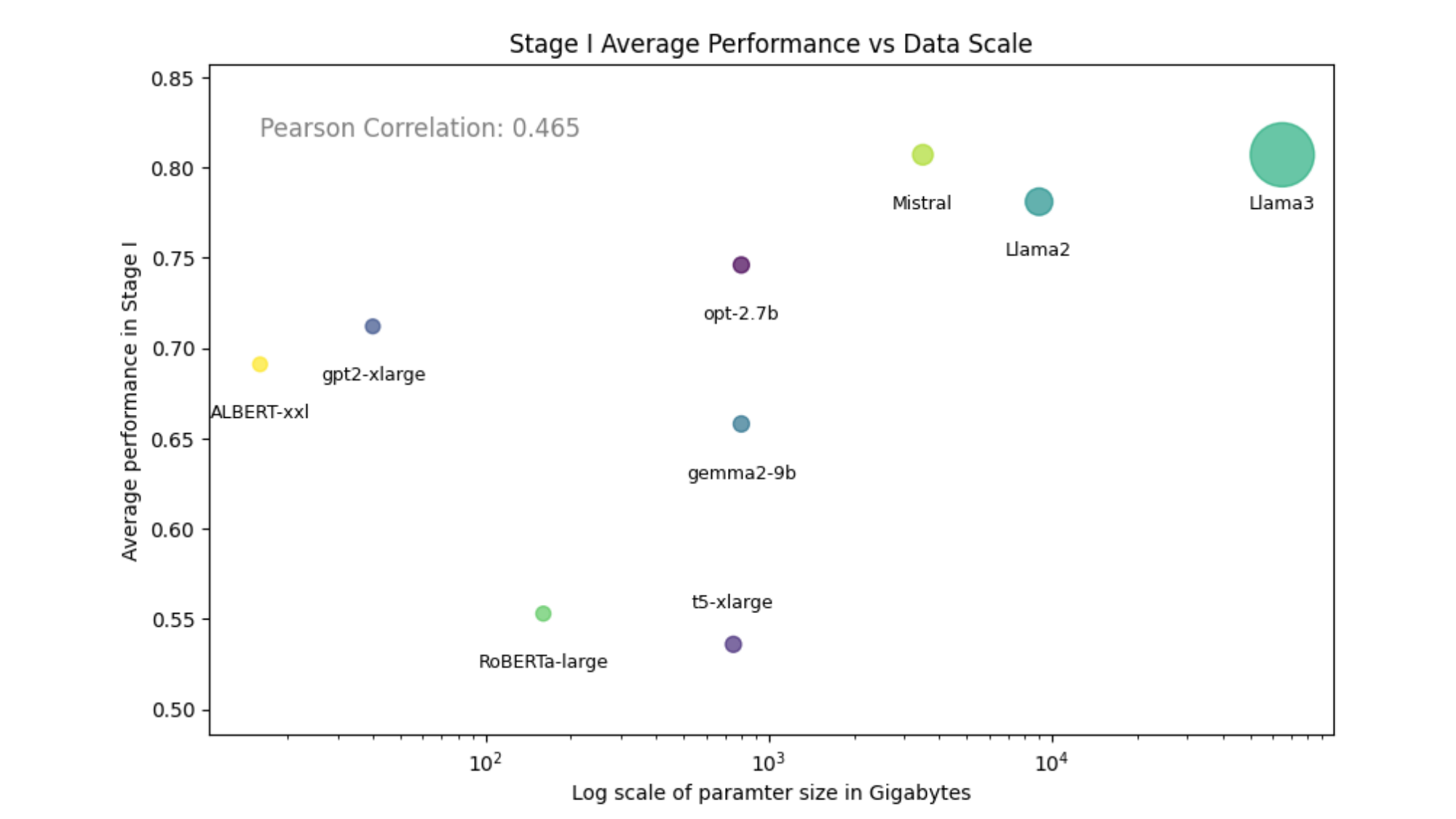}
    \caption{Stage I performance (normalized) vs. their data scale in the logarithm of Gigabyte.}
    \label{fig:stageIdata}
\end{figure*}

\begin{figure*}[h]
    \centering
    \includegraphics[width=\linewidth]{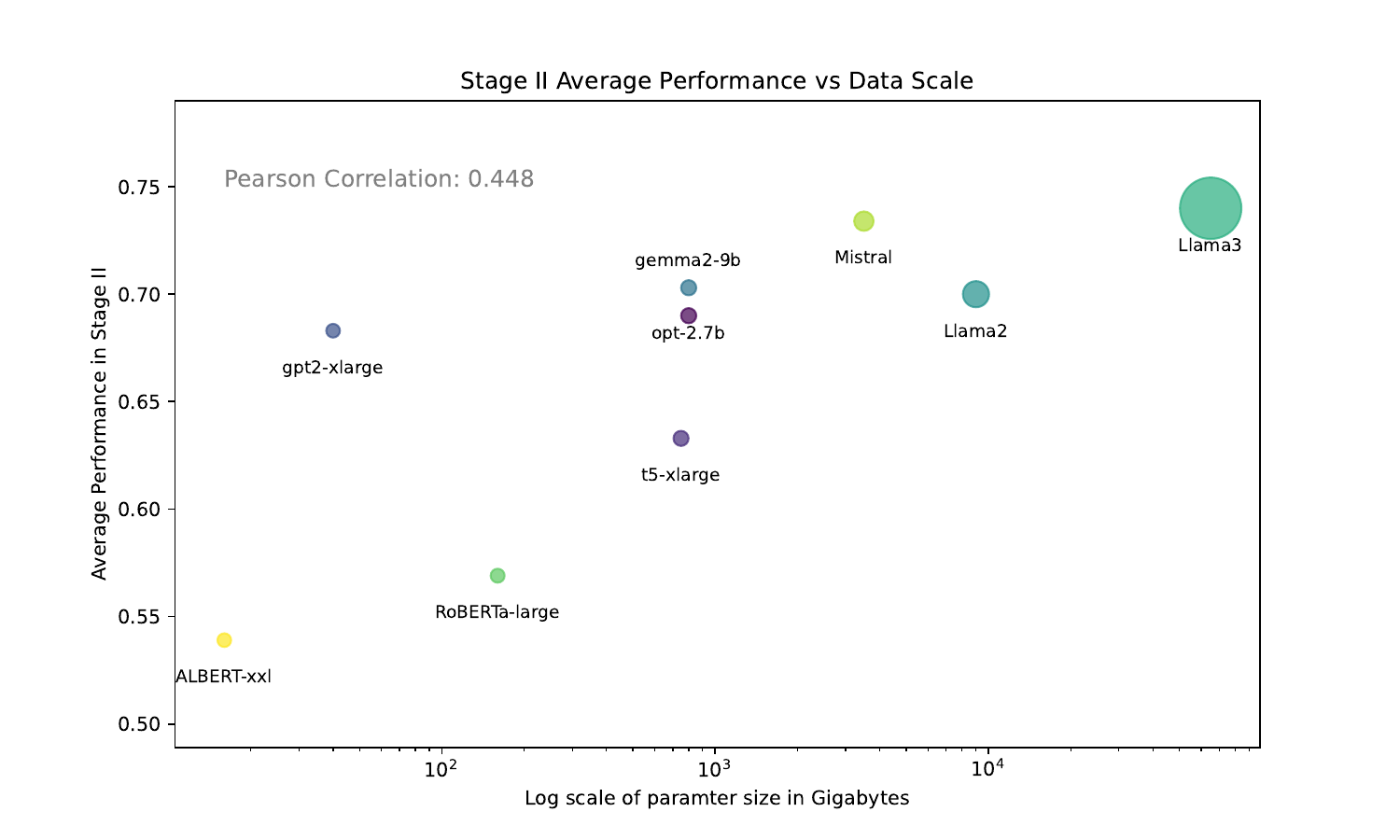}
    \caption{Stage II performance (normalized) vs. their data scale in the logarithm of Gigabyte.}
    \label{fig:stageIIdata}
\end{figure*}

\renewcommand{\arraystretch}{0.75} 
\begin{figure*}[h]
    \centering
    \includegraphics[width=\linewidth]{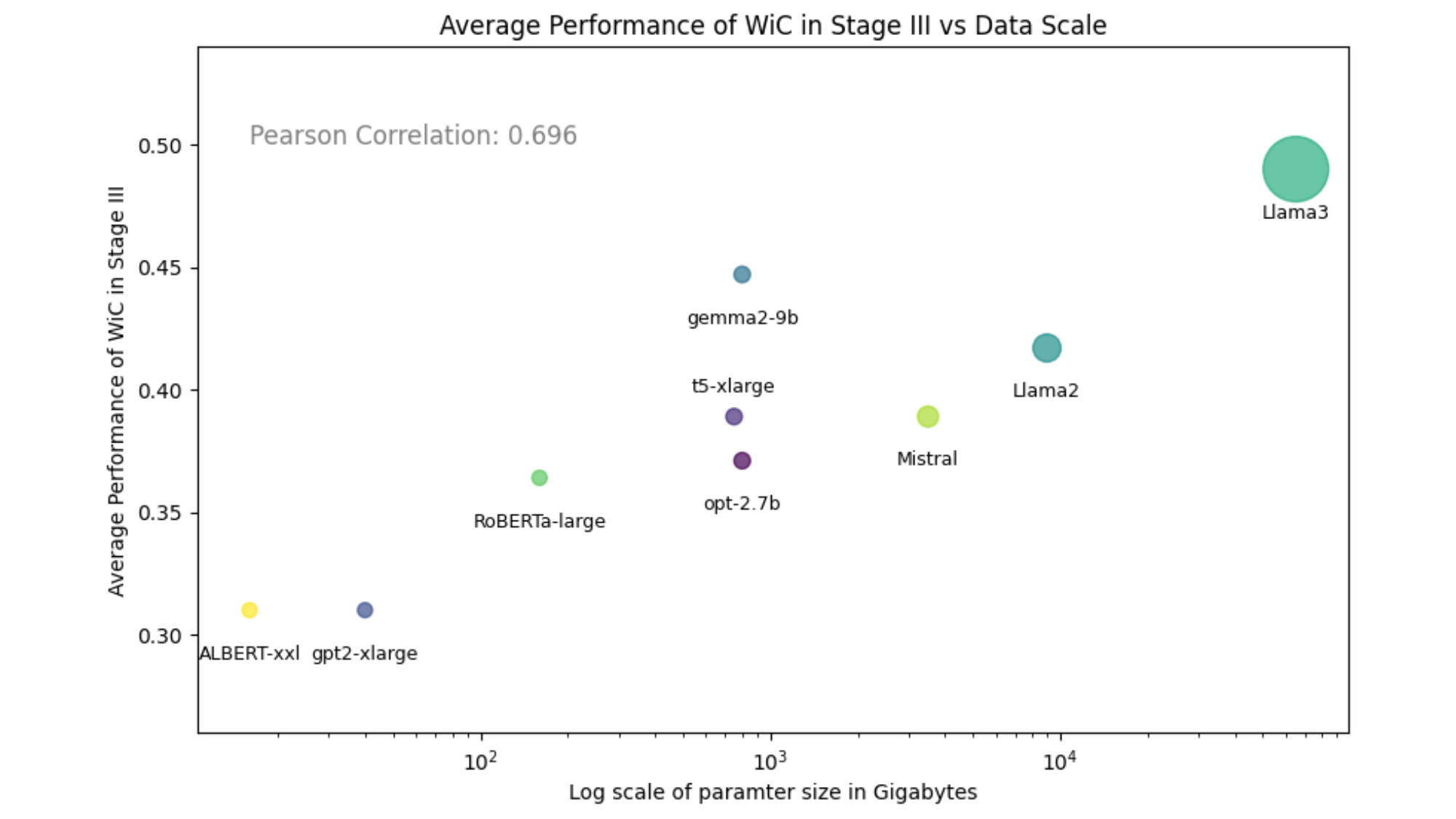}
    \caption{WiC in Stage III performance (normalized) vs. their data scale in the logarithm of Gigabyte.}
    \label{fig:stageIIIdata}
\end{figure*}

\begin{figure*}[h]
    \centering
    \includegraphics[width=\linewidth]{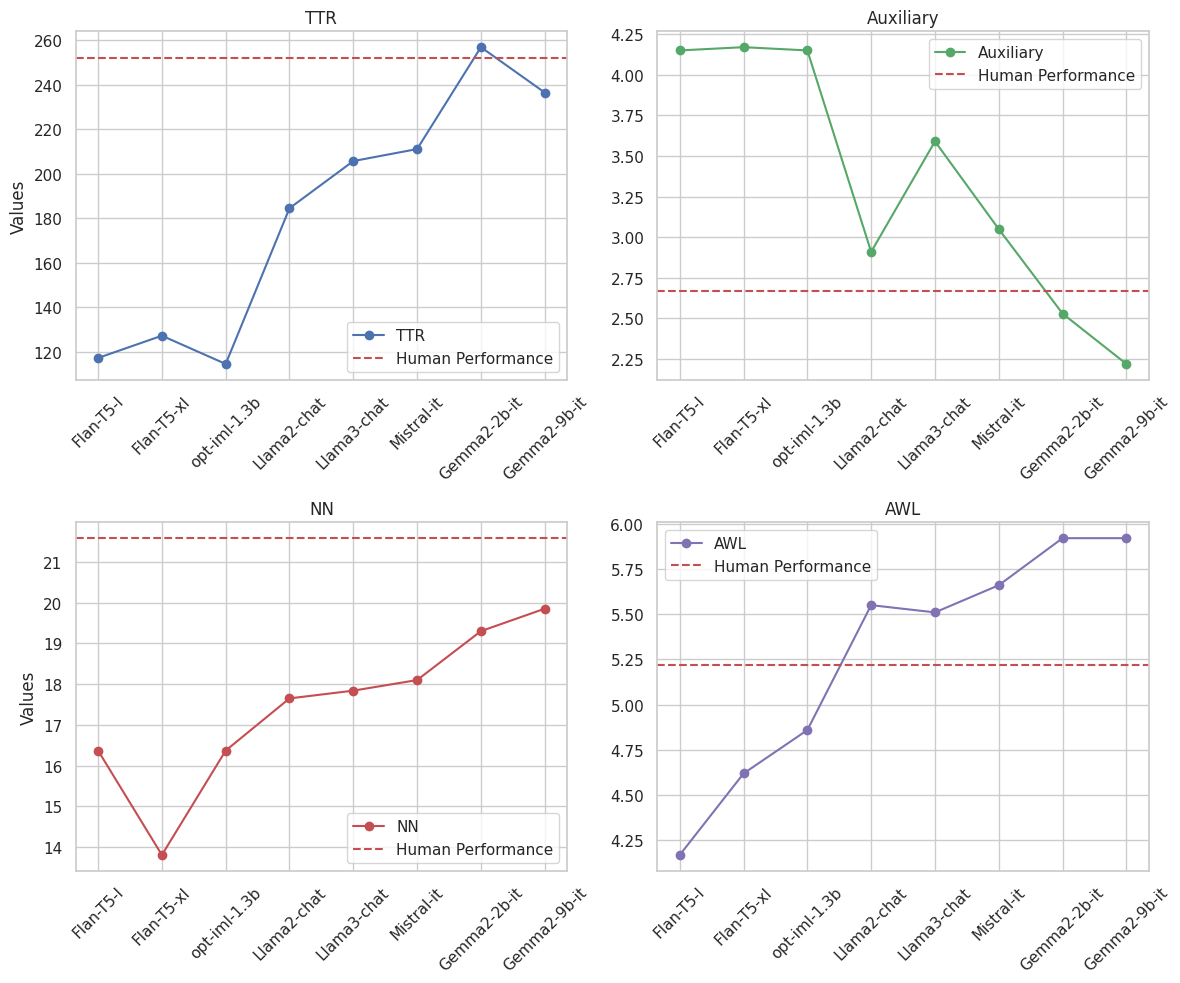}
    \caption{Four types of grammar metrics. Models are ordered by time}
    \label{fig:generation}
\end{figure*}

\begin{figure*}[h]
    \centering
    \includegraphics[width=\linewidth]{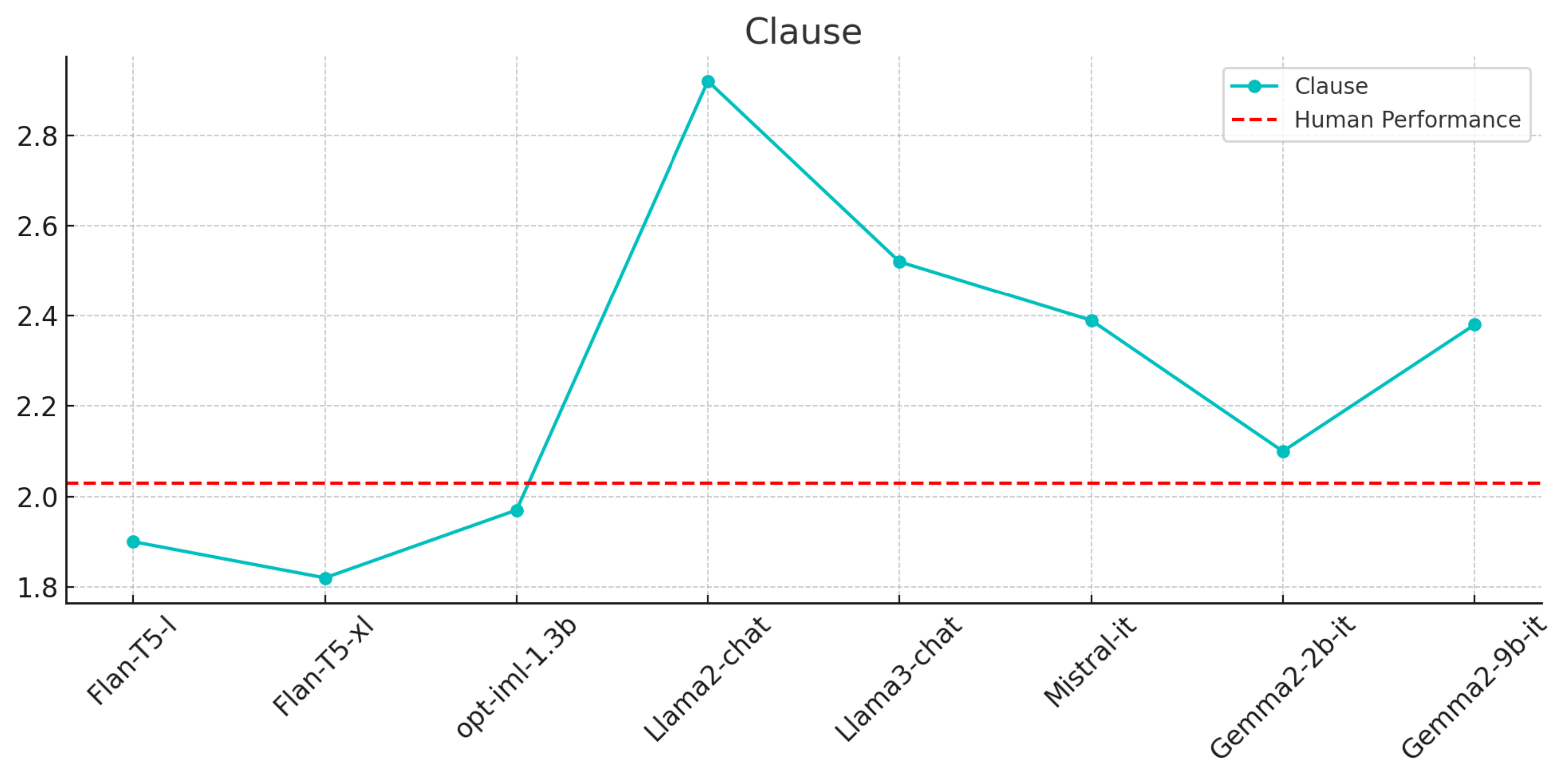}
    \caption{Clause grammar metrics. Models are ordered by time}
\end{figure*}


\begin{table*}[!t]
\centering
\begin{tabular}{lccc}
\toprule
Models & Raw Accuracy & 1-shot ICL & 0-shot CoT \\
\midrule
opt-iml-1.3b & \result{0.31} & \result{0.32\textcolor{green}{\small{ +0.06}}} & \result{0.32\textcolor{green}{\small{ +0.06}}} \\
Flan-t5-l & \result{0.42} & \result{0.38\textcolor{red}{\small{ -0.05}}} & \result{0.42\textcolor{gray}{\small{ +0.00}}} \\
Flan-t5-xl & \result{0.55} & \result{0.55\textcolor{gray}{\small{ +0.00}}} & \result{0.54\textcolor{gray}{\small{ -0.00}}} \\
Gemma2-2b-it & \result{0.49} & \result{0.46\textcolor{red}{\small{ -0.03}}} & \result{0.49\textcolor{gray}{\small{ +0.00}}} \\
Gemma2-9b-it & \result{0.72} & \result{0.76\textcolor{green}{\small{ +0.04}}} & \result{0.71\textcolor{red}{\small{ -0.01}}} \\
Llama2-7b-chat & \result{0.37} & \result{0.36\textcolor{red}{\small{ -0.01}}} & \result{0.36\textcolor{red}{\small{ -0.01}}} \\
Llama3-8b-chat & \result{0.58} & \result{0.56\textcolor{red}{\small{ -0.03}}} & \result{0.43\textcolor{red}{\small{ -0.15}}} \\
Mistral-7b-it & \result{0.55} & \result{0.55\textcolor{gray}{\small{ +0.00}}} & \result{0.53\textcolor{red}{\small{ -0.02}}} \\
\bottomrule
\end{tabular}
\caption{Model Performance with raw accuracy on ReClor Dataset with 1-shot ICL and 0-shot CoT.}
\label{tab:reclorfull}
\end{table*}

\begin{table*}[!t]
\centering
\begin{tabular}{lcccc}
\toprule
Grammar Phenomena & RoBERTa-base & T5-l & Gemma2-9b & Human\\
\midrule
passive\_2 & 0.60 & \textbf{0.87} & 0.75 & 0.86\\
determiner\_noun\_agreement\_with\_adj\_irregular\_1 & 0.50 & 0.83 & \textbf{0.89} & 0.94 \\
superlative\_quantifiers\_2 & \textbf{0.89} & 0.76 & 0.71 & 0.85 \\
wh\_questions\_subject\_gap\_long\_distance & 0.72 & \textbf{0.90} &0.80 & 0.85 \\
superlative\_quantifiers\_1 & 0.42 & \textbf{1.00} & 0.71& 0.94 \\
causative & 0.72 & \textbf{0.78} & 0.65 & 0.98 \\
\bottomrule
\end{tabular}
\caption{Selected results from BLiMP-comp of detailed grammar phenomena. We could notice the discrepancy in performance among the three models in these tasks, while humans could maintain high performance relatively.}
\label{tab:blimp-comp}
\end{table*}

\begin{table*}
\begin{tabular}{p{16cm}}
\subsubsection*{Examples of each task}
\\
\hline
\paragraph{one-word understanding}\\\\
\textbf{Question:} When you say something to someone’s ear quietly and secretly, you \_\_\_\_\_\_.  \\\\
\textbf{A)} repeat\\
\textbf{B)} whisper\\
\textbf{C)} discuss\\
\textbf{D)} cry\\
\textbf{Correct Answer: } B \\
\hline
\paragraph{Agent-Action-Object (AAO)} \\\\
\textbf{sentence\_good:} Tanya conceals Adam. \\ \textbf{sentence\_bad:} This ice cream conceals Adam. \\
\hline
\paragraph{bc-if-why}\\\\
\textbf{Premise:} If we keep up, they'll route. \\\textbf{Hypothesis: } They'll route if we keep up.\\
\textbf{Label:} Entailment\\
\hline
\paragraph{grammar-comp}\\\\
\textbf{Premise:} For Master P, neither is an appealing prospect. \\
\textbf{Hypothesis: } Master P found both projects to be appealing. \\
\textbf{Label:} Contradiction\\
\hline
\paragraph{CoLA}\\\\
\textbf{sentence:} The in loved peanut butter cookies.\\
\textbf{Label:} 0 (False)
\\
\hline
\paragraph{BLiMP-comp:} determiner\_noun\_agreement\_adj\_2\\\\
\textbf{sentence\_good:} Cynthia scans these hard books. \\
\textbf{sentence\_bad:} Cynthia scans this hard books.
\\
\hline
\paragraph{WiC}\\\\
\textbf{word:} carry\\
\textbf{sentence1:} You must carry your camping gear.\\
\textbf{sentence2:} Sound carries well over water.\\
\textbf{Label:} F (False)\\
\hline
\paragraph{ReClor}\\\\
\textbf{Context:} In a business whose owners and employees all belong to one family, the employees can be paid exceptionally low wages. Hence, general operating expenses are much lower than they would be for other business ventures, making profits higher. So a family business is a family's surest road to financial prosperity. \\\\
\textbf{Question:} The reasoning in the argument is flawed because the argument\\\\
\textbf{A)} ignores the fact that in a family business, paying family members low wages may itself reduce the family's prosperity
\\
\textbf{B)} presumes, without providing justification, that family members are willing to work for low wages in a family business because they believe that doing so promotes the family's prosperity
\\
\textbf{C)} ignores the fact that businesses that achieve high levels of customer satisfaction are often profitable even if they pay high wages 
\\
\textbf{D)} presumes, without providing justification, that only businesses with low general operating expenses can succeed\\
\textbf{Answer: } A \\
\hline 
\end{tabular}
\caption{One example from each dataset.}
\label{tab:example}
\end{table*}

\end{document}